\documentclass[12pt]{article}
\usepackage{amsfonts}
\usepackage{latexsym}
\usepackage{amssymb}
\usepackage{amsmath}
\usepackage{amsthm}
\usepackage[mathscr]{eucal}
\usepackage{graphicx}
\usepackage{hyperref}
\usepackage{caption}
\usepackage{subcaption}

\newtheorem{Lemma}{Lemma}

\newtheorem{Theorem}[Lemma]{Theorem}

\newtheorem{Definition}{Definition}

\renewcommand{\qed}{\hfill{\ \ \rule{2mm}{2mm}} \vspace{0.2in}}

\newcommand{\ind}{1\hspace{-2.3mm}{1}}

\begin{document}

\title{A Probabilistic Model for Data Redundancy in the Feature Domain}

\author{ \textbf{Ghurumuruhan Ganesan}
\thanks{E-Mail: \texttt{gganesan82@gmail.com} } \\
\ \\
IISER Bhopal}


\date{}
\maketitle

\begin{abstract}
In this paper, we use a probabilistic model to estimate the number of uncorrelated features in a large dataset. Our model allows for both pairwise feature correlation (collinearity) and interdependency of multiple features (multicollinearity) and we use the probabilistic method to obtain  upper and lower bounds of the same order, for the size of a feature set that exhibits low collinearity and low multicollinearity. We also prove an auxiliary result regarding mutually good constrained sets that is of independent interest.




\vspace{0.1in} \noindent \textbf{Key words:} Data Redundancy, Feature Domain, Probabilistic Model, Mutually Good Constrained Sets.

\vspace{0.1in} \noindent \textbf{AMS 2000 Subject Classification:} Primary: 60K35, 60J10;
\end{abstract}

\bigskip

\setcounter{equation}{0}
\renewcommand\theequation{\thesection.\arabic{equation}}
\section{Introduction}\label{intro}
The feature selection problem is a very important part of data preprocessing that crucially affects the overall performance in predictive analysis~\cite{kuhn}. Given a large dataset, statistical tests are typically performed to estimate the correlation between pairs and subsets of features and a subset of the total feature set is then chosen using standard feature selection methods like filters and wrappers~\cite{guyon}~\cite{wah}~\cite{yang}. This is done to reduce data redundancy and also improve the performance of the statistical or machine learning methodology to which the resulting data is fed~\cite{yu}.


In this paper, we use a probabilistic approach to the data feature redundancy problem by defining a random graph model that allows for both collinearity and multicolllinearity among features. We use an auxiliary result regarding the size of mutually good constrained  sets to obtain a lower bound on the minimum size of a feature set that has low collinearity and low multicollinearity.


In the following section, we state and prove our main result regarding the size of feature sets with low collinearity and multicollinearity, using mutually good constrained sets. We also prove a Lemma regarding the size of mutually good constrained sets, that is of independent interest.

\setcounter{equation}{0}
\renewcommand\theequation{\thesection.\arabic{equation}}
\section{Feature Domain Redundancy}\label{sec_feat}
In this Section,  we study the data redundancy problem from the feature domain perspective where we seek a subset of data features that are nearly uncorrelated with each other. To motivate the problem, suppose~\(W_i = (W_i(1),\ldots,W_i(m)), i \geq 1\) are independent and identically distributed (i.i.d.) elements belonging to some space~\({\cal W}.\)  We refer to~\(W_i\) as the~\(i^{th}\) data point and~\(W_i(j)\) as the~\(j^{th}\) \emph{feature} of the~\(i^{th}\) data point.

In general, the~\(m\) features in the dataset~\(\{W_i\}\) may be correlated with each other; i.e.~\(W_i(j)\) is not necessarily independent of~\(W_i(k)\) for~\(j \neq k\) and so statistical tests~\cite{kuhn} are performed to obtain estimates for the correlation between distinct pairs of features.  Using these estimates, we are interested  in determining a  ``nice" subset~\({\cal S} \subset \{1,2,\ldots,m\}\) of nearly uncorrelated features.

One heuristic method (see Chapter~\(3,\) pp.~\(47,\)~\cite{kuhn}) is to remove the minimum number of features iteratively, in such a way that all \emph{pairwise correlations} (also known as collinearity) of the remaining features are below a predetermined threshold. It is also possible that the dataset exhibits \emph{multicollinearity} where multiple features are interdependent on each other and  in our main result of this section, we use a probabilistic model to obtain high probability bounds for the minimum size of a nice  feature set with low collinearity and low multicollinearity.

We begin with a couple of definitions. Let~\(K_m\) be the complete graph on~\(m\) vertices and let~\(\{X(h)\}_{h \in K_m}\) be i.i.d.\ Bernoulli random variables satisfying
\[\mathbb{P}(X(h)=1) = p =1-\mathbb{P}(X(h)=0).\] Let~\(\{{\cal T}(v)\}_{1 \leq v \leq n}\) be random subsets of~\(\{1,2,\ldots,m\}\) that possibly depend on~\(\{X(.)\}.\) We assume that the sets~\({\cal T}(.)\) are consistent in the sense that~\(u \in {\cal T}(v)\) if and only if~\(v \in {\cal T}(u).\) We say that a set of vertices~\({\cal D}\) is \emph{nice} if:\\
\((i)\) For any~\(u,v \in {\cal D}\) we have~\(X(u,v) =0\) and\\
\((ii)\) There does not exist~\(u,v \in {\cal D}\) such that~\(u \in {\cal T}(v)\) (or~\(v \in {\cal T}(u)\)).


Letting~\(N_m\) be the largest size of a nice subset of~\(\{1,2,\ldots,m\},\) we have the following result.
\begin{Theorem}\label{red_free} For every~\(\gamma>0\) we have that
\begin{equation}\label{up_bound}
\mathbb{P}\left(N_m \leq (2+\gamma)\frac{\log{m}}{|\log(1-p)|}\right) \geq 1- \frac{1}{m^{\gamma}}.
\end{equation}
Conversely if~\(\max_{v} {\cal T}(v) \geq 1,\)
\begin{equation}\label{p_cond_axr}
p \geq \frac{1}{m^{\delta_1}} \text{ and } \tau := \mathbb{E}\max_{v} \#{\cal T}(v) \leq m^{\delta_2}
\end{equation}
for some constants~\(0< \delta_1,\delta_2<1,\) satisfying~\(0 < 2\delta_1+\delta_2 < 1,\) then there is a constant~\(\delta > 0\) such that
\begin{equation}\label{low_bound}
\mathbb{P}\left(N_m \geq (1-2\delta)\frac{\log{m}}{|\log(1-p)|} - \frac{\log\left(4\tau/p\right)}{|\log(1-p)|}\right) \geq 1- \frac{1}{m^{\delta}}.
\end{equation}
\end{Theorem}
In the context of the feature subset problem discussed prior to the statement of Theorem~\ref{red_free}, we could interpret~\(p\) as probability that features~\(u\) and~\(v\) are correlated and the set~\({\cal T}(v)\) as a subset of features that exhibit multicollinearity together with the feature~\(v.\) For example,~\({\cal T}(v)\) could be a subset of the features that result in a variance inflation factor (VIF)~\cite{kuhn} greater than~\(\lambda_{mc}\) for the feature~\(v,\) where~\(\lambda_{mc} > 0\) is a predetermined threshold. We recall that VIF measures the extent to which a particular feature depends on a subset of features and for more details, we refer to Chapter~\(3,\)~\cite{kuhn}.



Below, we use the following deviation estimate regarding of sums of independent Bernoulli random variables. Let~\(\{W_j\}_{1 \leq j \leq r}\) be independent Bernoulli random variables satisfying~\(\mathbb{P}(W_j = 1) = 1-\mathbb{P}(W_j = 0) > 0.\) If~\(S_r := \sum_{j=1}^{r} W_j, \theta_r := \mathbb{E}S_r\) and~\(0 < \gamma \leq \frac{1}{2},\) then
\begin{equation}\label{conc_est_f}
\mathbb{P}\left(\left|S_r - \theta_r\right| \geq \theta_r \gamma \right) \leq 2\exp\left(-\frac{\gamma^2}{4}\theta_r\right)
\end{equation}
for all \(r \geq 1.\) For a proof of~(\ref{conc_est_f}), we refer  to Corollary~\(A.1.14,\) pp.~\(312\)  of~\cite{alon}.

\emph{Proof of Theorem~\ref{red_free}}: We begin with the upper bound for~\(N_m.\) Let~\(G\) be the random subgraph of~\(K_m\) obtained by retaining all edges~\((u,v)\) satisfying~\(X(u,v) =1.\) The probability that the vertices~\(\{1,2,\ldots,T\}\) form a stable set in~\(G\)  (i.e. a set of vertices no two of which are adjacent in~\(G\)) is~\((1-p)^{T \choose 2}\) and so the probability that there exists a stable set of size at least~\(T\) in~\(G\) is bounded above by
\[m^{T} \cdot (1-p)^{T(T-1)/2}  =\exp\left(-T \left(\frac{T-1}{2}|\log\left(1-p\right)| - \log{m}\right)\right) \leq \frac{1}{m^{\gamma}}\] provided~\(T \geq 1 + (2+\gamma) \frac{\log{m}}{|\log(1-p)|}.\) This obtains the upper bound for~\(N_m\) in~(\ref{up_bound}).



In what follows, we obtain a lower bound for~\(N_m\) using an estimate for the size of \emph{mutually good constrained sets} derived in Lemma~\ref{good_thm_main} at the end of this section. We define the event~\(E_{tau} := \{\max_{v} \#{\cal T}(v) \leq \tau\cdot m^{\gamma}\}\) where~\(\gamma > 0\) is a constant to be determined later. From the Markov inequality we see that
\begin{equation}\label{e_tau}
\mathbb{P}(E_{tau}) \geq 1- \frac{1}{m^{\gamma}}
\end{equation}
and we henceforth assume that~\(E_{tau}\) occurs.

Next, we define the goodness function~\(f({\cal S})\) to be the set of all vertices not adjacent to any vertex of~\({\cal S}\) in~\(G\) and set the constraint function~\(g\)  as
\begin{equation}\label{g_def_two}
g(x,{\cal I}) =
\left\{
\begin{array}{ll}
1 &\;\;x \in \bigcup_{v \in {\cal I}} \left({\cal T}(v) \cup \{v\}\right) \\
0 &\;\;\text{otherwise},
\end{array}
\right.
\end{equation}
with~\({\cal E} = \{0,1\}\) and~\({\cal B} = \{1\}.\) The constraint~\(g\) ensures that we ``add" a new vertex in each iteration that is not adjacent to any of the previously added vertices and also does not belong to the ``conflict" set~\({\cal T}(v)\) of a previously added vertex~\(v.\)

Let~\(1  \leq  L \leq \frac{m}{2}\) be an integer to be determined later. Since~\(E_{tau}\) occurs, each~\({\cal T}(v)\) has size at most~\(\tau \cdot m^{\gamma}\) and so the parameter~\(q_i\) defined in~(\ref{ti_def}) is bounded above as
\begin{equation}\label{q_def}
q_{i} \leq \frac{i\tau \cdot m^{\gamma}}{m} \leq \frac{L\tau}{m^{1-\gamma}}.
\end{equation}
To estimate the term~\(p_i\) in~(\ref{pi_def}), we let~\({\cal S} = \{x_1,\ldots,x_{i}\}\)  be any deterministic set of~\(i\) vertices. A vertex~\(u \notin \{x_1,\ldots,x_{i}\}\) is good (i.e. not adjacent to any vertex of~\({\cal S}\) in~\(G\)) with probability~\((1-p)^{i}\) and so the expected number of vertices that are good with respect to~\({\cal S}\) is at least~\((m-i)(1-p)^{i} \geq \frac{m(1-p)^{i}}{2}.\)

By the standard deviation estimate~(\ref{conc_est_f}), we therefore get that the set~\(f({\cal S})\) of good vertices with respect to~\({\cal S}\) has size at least~\(\frac{1}{4} \cdot m(1-p)^{i}\) with probability at least~\[1-e^{-C_2m(1-p)^{i}} \geq 1-e^{-C_2m(1-p)^{L}}\] for some constant~\(C_2 > 0.\)
Therefore considering all possible choices of~\({\cal S}\) with~\(i \leq L-1\) vertices, we get that the fraction
\begin{equation}\label{pi_def2}
p_{L-1} \geq \min_{{\cal S}}\frac{f({\cal S})}{m} \geq \frac{(1-p)^{L}}{4}
\end{equation}
with probability at least~\(1-\zeta\) where
\begin{eqnarray}
\zeta &:=& \sum_{i=1}^{L-1} {m \choose i}e^{-C_2 m(1-p)^{L}} \nonumber\\
&\leq& L {m \choose L-1} \cdot e^{-C_2m(1-p)^{L}} \nonumber\\
&\leq& L m^{L-1} \cdot e^{-C_2m(1-p)^{L}}, \label{good_est}
\end{eqnarray}
by the unimodality of the Binomial coefficient for~\(L \leq \frac{m}{2}.\)



From the condition~\(p_{L-1} > q_{L-1}\) in Lemma~\ref{good_thm_main} and the estimates for~\(p_{L-1}\) and~\(q_{L-1}\) in~(\ref{pi_def2}) and~(\ref{q_def}) respectively, we get that if
\begin{equation}\label{theodora}
\frac{(1-p)^{L}}{4} > \frac{L\tau}{m^{1-\gamma}}
\end{equation}
then there exists a nice set of size~\(L\) in~\(G.\) Setting~\(L = \min\left(1,\theta\frac{\log{m}}{|\log(1-p)|}\right)\) with~\(0 < \theta  <1\) and using the inequality~\(|\log(1-p)| > p,\) we see that~(\ref{theodora}) is true if
\[\frac{1}{4m^{\theta}} > \frac{\theta \tau \log{m}}{pm^{1-\gamma}}\]
or  equivalently if
\[\log{\theta} + \theta \log{m} < \log\left(\frac{p}{4\tau}\right) + (1-\gamma)\log{m} - \log\log{m}.\]
We set
\[\theta  := 1-2\gamma + \frac{\log\left(p/4\tau\right)}{\log{m}}\] where~\(\gamma >0\) is chosen such that~\(\delta_1 < 2\gamma < 1-\delta_1-\delta_2.\) This is possible by Theorem statement. Using the condition~\(p \geq \frac{1}{m^{\delta_1}}, \tau \leq m^{\delta_2}\) and the fact that~\(\delta_1+\delta_2 < 1\) strictly (see Theorem statement), we get that~\(\theta > 0\) strictly and moreover,
\begin{equation}\label{mp_est}
m(1-p)^{L} = m^{1-\theta} = m^{2\gamma} \cdot \frac{4\tau}{p}  \geq m^{2\gamma}
\end{equation}
since~\(p < 1\) and~\(\tau \geq 1,\) again by Theorem statement.

Also
\begin{equation}\label{theo_two}
L \leq \frac{\log{m}}{|\log(1-p)|} < \frac{\log{m}}{p} < m^{\delta_1}\log{m}
\end{equation}
and so plugging~(\ref{theo_two}) and~(\ref{mp_est})  into~(\ref{good_est}), we get
\[\zeta \leq m^{\delta_1} \cdot \log{m} \cdot \exp\left(m^{\delta_1} (\log{m})^2\right) \cdot \exp\left(-C_2m^{2\gamma}\right) \longrightarrow 0,\] by our choice of~\(2\gamma > \delta_1.\) Combining the estimate~(\ref{e_tau}) for the event~\(E_{tau}\) and the estimate~(\ref{good_est}), we therefore get the lower bound in~(\ref{low_bound}) and this completes the proof of the Theorem.~\(\qed\)


\subsection*{Mutually Good Constrained Sets}
Let~\({\cal U}\) be a finite set containing~\(N\) elements and~\(2^{{\cal U}}\) be the set of all subsets of~\({\cal U}.\) We have the following definition.
\begin{Definition}\label{good_def}
A map~\(f :2^{{\cal U}} \rightarrow 2^{{\cal U}}\) is said to be a \emph{goodness function} if for any two sets~\({\cal S}_1,{\cal S}_2\subseteq {\cal U}\) we have:\\
\((i)\) The set~\({\cal S}_1 \subseteq f({\cal S}_2)\) if and only if~\({\cal S}_2 \subseteq f({\cal S}_1).\)\\
\((ii)\) The set~\(f({\cal S}_1 \cup {\cal S}_2) = f({\cal S}_1) \cap f({\cal S}_2).\)
\end{Definition}
We use the notation~\(f(\emptyset) = {\cal U}\) and say that~\(f({\cal S})\) is the set of elements that are~\(f-\)good or simply good with respect to~\({\cal S}.\) A set of elements~\({\cal S} \subseteq {\cal U}\) is said to be \emph{mutually good} if for any~\({\cal I} \subset {\cal S},\) we have that~\({\cal S} \setminus {\cal I} \subseteq f({\cal I}).\)

For example, if~\({\cal U}\) is the set of vertices in a graph, then the function~\(f_0({\cal S})\) that determines the set of all vertices not adjacent to any vertex of~\({\cal S}\) is a goodness function. A stable set, i.e. a set of vertices no two of which are adjacent to each other, is a mutually good set with respect to the goodness function~\(f_0.\)


For a set~\({\cal E},\) a~\(({\cal U},{\cal E})-\)constraint or simply a constraint is a map \[g : {\cal U} \times 2^{{\cal  U}}~\rightarrow~{\cal E}.\] For sets~\({\cal I} \subseteq {\cal U}\) and~\({\cal B} \subseteq {\cal E},\) we say that~\(x \in {\cal U}\) satisfies the~\({\cal B}-\)constraint with respect to~\({\cal I}\) if~\(g(x,{\cal I}) \in {\cal B}.\) We also say that~\({\cal I}\) is a~\({\cal B}-\)\emph{constrained} set if each~\(y \in {\cal I}\) satisfies the~\({\cal B}-\)constraint with respect to~\({\cal I}\setminus \{y\}.\) Finally, we define~\[h({\cal I}) := \{x \in {\cal U}: g(x,{\cal I}) \notin {\cal B}\}\] to be the set of all elements that do not satisfy the~\({\cal B}-\)constraint with respect to~\({\cal I}.\)

Continuing with the graph example, let~\({\cal E} = \{0,1\}\) and~\({\cal B}=\{1\}.\) The map~\(g_0(x,{\cal I})\) which equals~\(1\) if~\(x\) is not adjacent to any vertex of~\({\cal I}\) and zero otherwise, is an example of a~\({\cal B}-\)constraint. Any stable set~\({\cal I}\) is a constrained set and the set~\(h({\cal I})\) is the set of all vertices adjacent to some vertex in~\({\cal I}.\)




We have the following result regarding size of mutually good sets.
\begin{Lemma}\label{good_thm_main} For sets~\({\cal U}\) and~\({\cal E}\) let~\(f\) and~\(g\) be the goodness and constraint functions, respectively, as defined above and let~\({\cal B} \subseteq {\cal E}\) be any subset. For integer~\(1 \leq  i \leq N = \#{\cal U}\) let
\begin{equation}\label{pi_def}
p_i := \min_{{\cal I} \subseteq {\cal U} : \#{\cal I} \leq i} \frac{\#f({\cal I})}{N}
\end{equation}
be the minimum fraction of elements that are good with respect to~\({\cal B}-\)constrained sets of cardinality at most~\(i.\) Similarly, let
\begin{equation}\label{ti_def}
q_i := \max_{{\cal I} \subseteq {\cal U}: \#{\cal I} \leq i} \frac{\#h\left({\cal I}\right)}{N}
\end{equation}
be the maximum fraction of elements not satisfying the~\({\cal B}-\)constraint with respect to~\({\cal B}-\)constrained sets of cardinality at most~\(i.\)
If~\(p_{L-1} > q_{L-1},\) then there exists a mutually good~\({\cal B}-\)constrained set of cardinality~\(L.\)
\end{Lemma}
Any single set in~\({\cal U}\) is assumed to be a mutually good set and so we always set~\(p_1 = 1 = 1-q_1.\) In the expressions for~\(p_i\) and~\(q_i\) in~(\ref{pi_def}) and~(\ref{ti_def}), the minimum and maximum are respectively taken over all \emph{constrained} sets of size at most~\(i.\) Therefore a lower bound for~\(p_i\) and an upper bound for~\(q_i\) is simply obtained by considering the minimum and maximum, respectively, over \emph{all} sets (constrained or not) of cardinality at most~\(i.\)

As we see from the graph theory example above, conditions could sometimes be posed both as a goodness function or as a constraint function. We pick the condition occurring with the lowest probability as a goodness function and identify the rest as constraints. 


We now use the probabilistic method to prove Lemma~\ref{good_thm_main}.\\
\emph{Proof of Lemma~\ref{good_thm_main}}: Let~\(X_1,\ldots,X_L\) be independently and uniformly chosen from~\({\cal U}.\)  For~\(1 \leq i \leq L\) let~\(E_i\) be the event that~\(\{X_1,\ldots,X_i\}\) is a mutually good set and let~\(H_i\) be the event that~\(\{X_1,\ldots,X_i\}\) is a~\({\cal B}-\)constrained set
and set~\(J_i := E_i \cap H_i.\) Clearly~\(J_{i+1} \subseteq J_{i}\) for~\(1 \leq i \leq L-1\) and suppose that the event~\(J_{L-1}\) occurs.
Given~\({\cal F}_{L-1} := \{X_1,\ldots,X_{L-1}\}\) we have that\\\(X_L \in f(\{X_1,\ldots,X_{L-1}\})\) with probability
\begin{equation}\label{plb}
\frac{\#f(\{X_1,\ldots,X_{L-1}\})}{N} \geq p_{L-1}
\end{equation}
since~\(\{X_1,\ldots,X_{L-1}\}\) is known to be a~\({\cal B}-\)constrained set, due to the occurrence of the event~\(J_{L-1}.\)

Again due to the event~\(J_{L-1},\) we know that~\(\{X_1,\ldots,X_{L-1}\}\) is also a mutually good set. We now use the properties~\((i)-(ii)\) in the Definition~\ref{good_def} to show that if~\(X_L \in f(\{X_1,\ldots,X_{L-1}\}),\) then~\({\cal S} := \{X_1,\ldots,X_L\}\) is a mutually good set as well. Indeed, let~\({\cal I} \subseteq \{X_1,\ldots,X_L\}\) be any set. If~\(X_L \in {\cal S} \setminus {\cal I},\) then
\begin{equation}\label{xla}
X_L \in f(\{X_1,\ldots,X_L\}) \subseteq f({\cal I})
\end{equation}
by property~\((ii)\) in Definition~\ref{good_def}. By the mutual goodness of~\(\{X_1,\ldots,X_{L-1}\},\) we already have that
\begin{equation}\label{xlb}
{\cal S} \setminus \left({\cal I} \cup \{X_L\}\right) \subseteq f({\cal I})
\end{equation}
and so combining~(\ref{xla}) and~(\ref{xlb}) we get~\({\cal S} \setminus {\cal I}  \subseteq f({\cal I}).\)

On the other hand if~\(X_L \in {\cal I},\) then using~\(X_L \in f(\{X_1,\ldots,X_{L-1}\}),\)
we get that
\begin{equation}\label{xlc}
{\cal S} \setminus {\cal I} \subseteq \{X_1,\ldots,X_{L-1}\} \subseteq f(\{X_L\})
\end{equation}
by property~\((i)\) in Definition~\ref{good_def}. As before, by the mutual goodness of the set\\\(\{X_1,\ldots,X_{L-1}\},\) we have that
\begin{equation}\label{xld}
{\cal S} \setminus {\cal I} \subseteq f({\cal I}\setminus \{X_L\})
\end{equation}
and so combining~(\ref{xlc}) and~(\ref{xld}) we get
\[{\cal S} \setminus {\cal I} \subseteq f({\cal I} \setminus \{X_L\}) \cap f(\{X_L\}) = f({\cal I})\]
by property~\((ii)\) in Definition~\ref{good_def}.

Summarizing we have that if~\(J_{L-1}\) occurs and~\(X_L \in f(\{X_1,\ldots,X_{L-1}\}),\) then~\(\{X_1,\ldots,X_L\}\) is a mutually good set and so from the probability estimate~(\ref{plb}), we get
\begin{equation}\label{pal}
\mathbb{P}(E_L \mid {\cal F}_{L-1}) \cdot \ind(J_{L-1}) \geq p_{L-1} \cdot \ind(J_{L-1}),
\end{equation}
where~\(\ind(.)\) refers to the indicator function. Similarly if the event~\(J_{L-1}\) occurs, then~\(\{X_1,\ldots,X_{L-1}\}\) is already a constrained set and so the probability that~\(\{X_L\}\) does not satisfy the~\({\cal B}-\)constraint with respect to~\(\{X_1,\ldots,X_{L-1}\}\) is at most~\(q_{L-1},\) by~(\ref{ti_def}). Consequently,
\begin{equation}\label{pbl}
\mathbb{P}(H^c_L \mid {\cal F}_{L-1}) \cdot \ind(J_{L-1}) \leq q_{L-1}\cdot \ind(J_{L-1}).
\end{equation}

Using~\(\mathbb{P}(A \cap B) \geq \mathbb{P}(A) - \mathbb{P}(B^c)\) with~\(A = E_L\) and~\(B = H_L,\) we get from~(\ref{pal}) and~(\ref{pbl}) that the conditional probability of both~\(E_L\) and~\(H_L\) happening is at least~\(p_{L-1}-q_{L-1}.\)
In other words,
\begin{eqnarray}
\mathbb{P}(J_L \mid {\cal F}_{L-1}) \cdot \ind(J_{L-1}) &=& \mathbb{P}(E_L \cap H_L \mid {\cal F}_{L-1}) \cdot \ind(J_{L-1}) \nonumber\\
&\geq& (p_{L-1}-q_{L-1})\cdot \ind(J_{L-1}). \label{pcL}
\end{eqnarray}
Taking expectations and using the fact that~\(J_L \subset J_{L-1}\) we get that
\[\mathbb{P}(J_L) \geq (p_{L-1}-q_{L-1}) \cdot \mathbb{P}(J_{L-1}).\]
Continuing iteratively, we get that
\begin{equation}\label{pc_L}
\mathbb{P}(J_L) \geq \prod_{j=2}^{L-1} (p_{j}-q_j) \cdot \mathbb{P}(J_1) = \prod_{j=2}^{L-1} (p_j-q_j) \cdot (1-q_1)
\end{equation}
since~\(p_1 = 1\) (see discussion following the statement of Lemma~\ref{good_thm_main}). By definition,~\(p_j\) as defined in~(\ref{pi_def}) is decreasing in~\(j\) and~\(q_j\) as defined in~(\ref{ti_def}) is increasing in~\(j.\)
Therefore if~\(p_{L-1} > q_{L-1},\) then we get that~\(\mathbb{P}(J_L) > 0\) and this proves the Lemma.~\(\qed\)

\emph{\underline{Acknowledgement}}: I thank Professors Rahul Roy, Thomas Mountford, Federico Camia, Alberto Gandolfi, Lasha Ephremidze and C. R. Subramanian for crucial comments and also thank IMSc and IISER Bhopal for my fellowships.



\bibliographystyle{plain}

\end{document}